\documentclass[letterpaper]{article} 
\usepackage{aaai2027}  
\usepackage[hyphens]{url}  
\usepackage{graphicx} 
\def\UrlFont{\rm}  
\usepackage{natbib}  
\usepackage{caption} 
\usepackage{algorithm}
\usepackage{algorithmic}

\usepackage{amsmath}
\usepackage{amssymb}
\usepackage{amsthm}

\theoremstyle{definition}
\newtheorem{definition}{Definition}
\newtheorem*{problem}{Problem Definition}

\usepackage{newfloat}
\usepackage{listings}
\DeclareCaptionStyle{ruled}{labelfont=normalfont,labelsep=colon,strut=off} 
\floatstyle{ruled}
\newfloat{listing}{tb}{lst}{}
\floatname{listing}{Listing}

\usepackage{booktabs}

\title{ORBITER: Conflict-Aware Decision-Making for Agentic Last-Mile Delivery}

\author{
    Mingzhao Li\textsuperscript{\rm 1},
    Chenxi Liu\textsuperscript{\rm 1},
    Yan Zhao\textsuperscript{\rm 2},
    Hao Miao\textsuperscript{\rm 3}
}
\affiliations{
    \textsuperscript{\rm 1}Centre for Artificial Intelligence and Robotics, Hong Kong Institute of Science \& Innovation, \\Chinese Academy of Sciences, Hong Kong SAR\\
    \textsuperscript{\rm 2}Shenzhen Institute for Advanced Study, \\University of Electronic Science and Technology of China, Shenzhen, China\\
    \textsuperscript{\rm 3}The Hong Kong Polytechnic University, Hong Kong SAR\\
    \{mingzhao.li,chenxi.liu\}@cair-cas.org.hk, zhaoyan@uestc.edu.cn, haomiao@polyu.edu.hk
}

\begin{document}

\maketitle

\begin{abstract}
Last-mile delivery aims to handle dynamically arriving orders with couriers while modeling complex spatial and temporal correlations. 
Recent learning-based methods model spatiotemporal dependencies among orders to predict courier service sequences, but leave next-order decision making unexplained. Describing the current delivery state in language allows LLMs to reason explicitly about the spatial, temporal, and behavioral cues behind an individual decision. As direct predictors, however, LLMs remain sensitive to task presentation and often produce unreliable decisions.
To address these challenges, we introduce \textbf{ORBITER}, an agentic \textbf{O}rder A\textbf{rbiter} for next-order decision-making in last-mile delivery.
ORBITER models courier service through decision points, each containing the courier's spatiotemporal state and visible orders and exposing local trade-offs for modeling and verification.
Fixed proposers rank the candidates, and a structured report identifies where their rankings disagree. The LLM uses task-specific tools to gather evidence on the leading alternatives, while an independent critic checks the resulting decision against that evidence.
We conduct extensive evaluations on data in four cities, where ORBITER outperforms existing state-of-the-art baselines by up to 9.2\% on average showing its effectiveness. 


\end{abstract}


\section{Introduction}


Last-mile delivery links logistics facilities to end customers and directly affects operational efficiency, courier experience, and customer satisfaction \cite{merchan20242021}. As e-commerce and on-demand services grow, platforms must coordinate rising order volumes across dispersed locations under tight deadlines and changing conditions \cite{olsson2019framework,boysen2021lastmile}.
These demands make last-mile delivery an important problem in logistics and spatiotemporal data mining.


Growing operational logs have made data-driven prediction central to last-mile service.
Sequence decoders and graph models capture dependencies among unfinished tasks \cite{wen2021deeproute,wen2022graph2route}, while related routing models use attention or graph neural networks to model candidate interactions \cite{kool2019attention,prates2019tsp}.
Reinforcement and imitation learning model routing policies and service behavior \cite{kwon2020pomo,feng2023ilroute}.
These methods predict service sequences, supporting downstream planning and time estimation \cite{pegadobardayo2024predictive}.
LLMs offer in-context adaptation and explicit reasoning \cite{brown2020language,kojima2022zeroshot}, together with tool use \cite{schick2023toolformer}.
Mobility predictors combine historical trajectories with temporal, semantic, social, and geographic context \cite{xue2021mobtcast}.
Recent LLM work represents trajectories and context through language inputs \cite{li2024llm4poi}, while agentic methods add personal memory and external tools \cite{feng2025agentmove,du2025trajagent}.
Trained predictors capture task patterns, whereas LLMs relate spatiotemporal and behavioral cues to context.

\begin{figure}[t]
\centering
\includegraphics[width=\columnwidth]{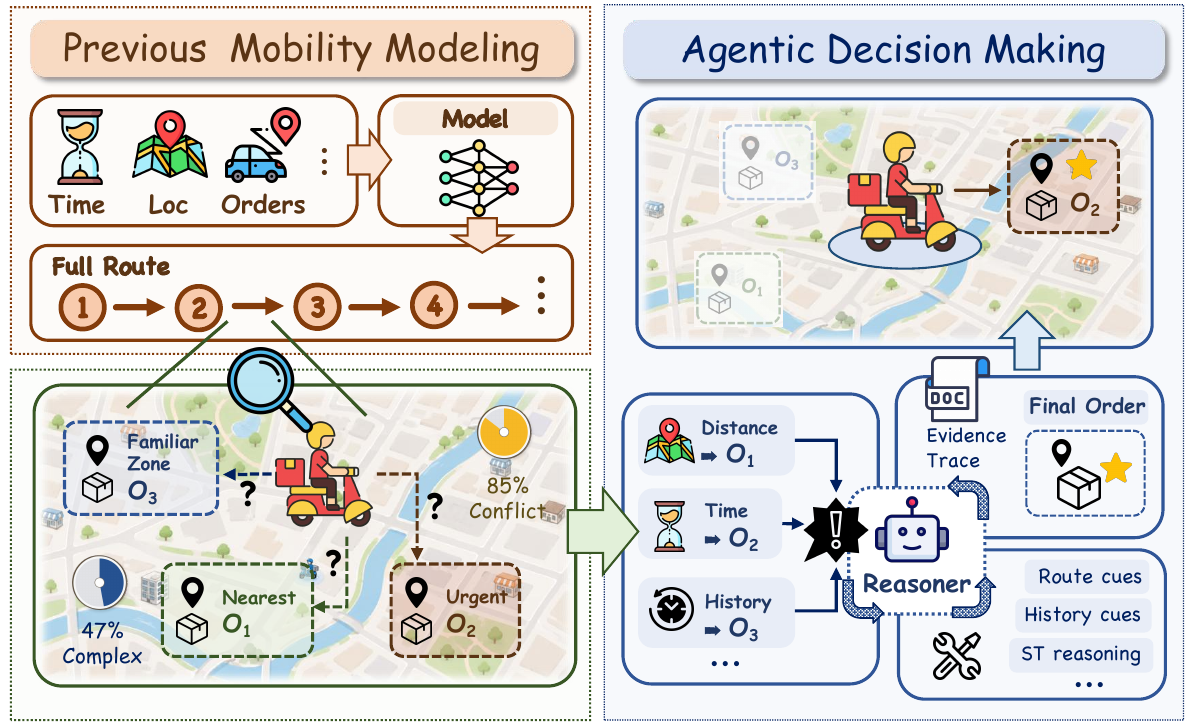}
\caption{Previous mobility modeling compresses local choices into full routes, whereas agentic decision making resolves conflicting cues with auditable evidence.}
\label{fig:motivation}
\end{figure}


However, last-mile service consists of local decisions under changing spatiotemporal constraints.
Deep models predict full service sequences, folding candidate trade-offs into the final route.
In dynamic pickup, orders arrive as the courier's location, waiting times, and deadline slack change \cite{wu2024lade}. As shown in the left panel of Figure~\ref{fig:motivation}, distance, deadlines, and service history may favor different candidates, yet end-to-end scores do not explain how conflicts are resolved. LLMs compare these factors but are often used as final predictors. Prompts, memory, retrieval, and tools add context, but irrelevant information can distract the model, making its output sensitive to presentation \cite{zheng2024picture}.
Tool outputs may not support the final decision, leaving explanations as unverified post-hoc justifications \cite{turpin2023unfaithful}.


These limitations leave three challenges for next-order decision making:
(1)~\textit{Decision granularity.}
Prior work operates at different granularities, from generating service sequences over available tasks \cite{wen2021deeproute,wen2022graph2route,liu2025mrgrp,rashidi2025practical} to classifying or ranking the next location among candidates \cite{luca2023overlap}.
These formulations optimize the final prediction, leaving comparisons among available orders and the local spatiotemporal cues behind each decision implicit.
The challenge is to represent each next-order decision together with the context in which it is made.
(2)~\textit{Behavioral complexity.}
A courier's next-order decision may depend on distance, deadlines, waiting time, and prior service patterns, which often favor different candidates.
In the Shanghai pickup records of LaDe, distance and deadline disagree in 55.4\% of decisions; after adding waiting time, 92.5\% have no order preferred by all three criteria.
No single operating rule can therefore account for courier behavior across such conflicts.
(3)~\textit{State-dependent conflict.}
Not all available information is equally useful for a particular choice. Evidence may rule out one candidate without separating those that remain, changing what needs to be examined next. Presenting all information at once can add irrelevant detail and prompt a decision before the central conflict is settled.


To address these challenges, we propose \textbf{ORBITER}, a conflict-aware agentic framework for spatiotemporal next-order decision making.
ORBITER first shifts the modeling unit from complete routes to individual next-order decisions, as shown on the right of Figure~\ref{fig:motivation}.
It generates a decision point after each completed service event from spatiotemporal logs, retaining the orders visible at that time while excluding information from later events.
All subsequent components operate on this inference-safe local state.
To examine behavioral trade-offs without requiring the LLM to search the full candidate space, ORBITER pairs an LLM agent with heterogeneous proposers.
The \emph{Heterogeneous Proposal Generator} uses a fixed panel of heuristic, statistical, and deep learning models to produce complementary candidate rankings.
A structured disagreement report combines their Top-1 votes, full rankings, and candidate attributes to identify model-supported candidates, the reference proposal and its leading rival, and the principal spatiotemporal conflict.
The agent receives this bounded hypothesis set, compares the relevant spatial, temporal, and historical cues, and calls task-specific tools to gather evidence.
For conflicts that remain unresolved, we introduce a hypothesis-verification loop in which each candidate becomes a testable next-order hypothesis.
Supporting and opposing evidence update its status, while the remaining conflict determines which comparison to examine next.
An independent critic reviews the proposal and evidence and may request an additional comparison.
The \emph{Evidence-Grounded Decision} stage then finalizes the next-order decision, retains the leading alternatives, and records its supporting evidence.


The main contributions are summarized as follows
\vspace{0.5cm}
\begin{itemize}
    \item We propose ORBITER, a conflict-aware agentic framework for spatiotemporal next-order decision making in last-mile delivery, combining the task-specific modeling of trained predictors with explicit reasoning by an LLM.
    \item We formulate last-mile service as a sequence of decision points, each preserving the local spatiotemporal state and available orders. This formulation makes individual courier decisions explicit for modeling and verification.
    \item We develop a Heterogeneous Proposal Generator and Conflict-Aware Agentic Reasoning for adjudicating model disagreements. A structured disagreement report narrows the competing hypotheses, while a hypothesis-verification loop gathers evidence as the unresolved conflict evolves.
    \item We conduct extensive evaluations on four city subsets of LaDe-P, where ORBITER consistently outperforms state-of-the-art baselines, with a 9.2\% average improvement in next-order decision accuracy.

\end{itemize}

\section{Related Work}

\textbf{Deep Learning Based Mobility Prediction.} Significant efforts have been made in mobility prediction using deep learning models, encompassing research from both location prediction and route prediction.
The former infers where an individual will go next from trajectories and spatiotemporal context, where deep sequential, attention, and graph models dominate
\cite{kong2018hstlstm,hong2022you,sun2020lstpm,luo2021stan,lin2021ctle,yang2022getnext,rao2022graphflashback}.
These methods profile general users over an open location set, while a courier picks from candidate orders fixed by the current task state.
The latter predicts the visiting order of unfinished delivery tasks \cite{pegadobardayo2024predictive}:
DeepRoute, Graph2Route, DRL4Route, and MRGRP advance it with pointer decoding, dynamic spatiotemporal graphs, reinforcement learning, and multi-relational graphs
\cite{wen2021deeproute,wen2022graph2route,mao2023drl4route,liu2025mrgrp}.
Both lines are end-to-end black-box mappings: they produce a result but cannot answer why the courier makes a particular choice at a decision point, so the prediction is neither interpretable nor auditable.
This motivates us to recast service as a sequence of real decision points and make explicit how each choice is formed, rather than merely fitting the outcomes.

\noindent\textbf{LLM-Based Mobility Reasoning.} Large language models have recently been brought into mobility modeling, which studies how individuals move through urban space, from ordinary users visiting the next place to couriers serving pickup and delivery orders.
These efforts differ in how much machinery is built around the model.
The lightest prompt it directly, casting stays and candidate POIs as text \cite{wang2023llmmob,feng2024llmmove}.
Others adapt or augment the model itself, fine-tuning it on trajectory prompts \cite{li2024llm4poi}, injecting geographic encodings \cite{liu2025gallm}, or retrieving similar users and trajectories as context \cite{wu2026mrpllm,li2025rallmpoi}.
The most elaborate are agentic: prediction is decomposed into memory and knowledge modules \cite{feng2025agentmove}, agents gain mobility tools or specialized small models \cite{chen2026agentmob,patil2024gorilla}, and the recipe reaches delivery route planning \cite{li2026talkingtrails}.
LLM agents also simulate behavior, acting as urban residents \cite{wang2024llmob} or delivery riders \cite{zhang2026llmdr}.
Across these designs, however, the LLM retains the same role: it acts as the predictor or simulator, without explicit checks against task evidence.
We flip this role.
Rather than adding another predictor, we direct the LLM's reasoning at the conflicts among heterogeneous predictors, so that disagreement is no longer noise but a signal to be adjudicated against auditable evidence within the candidate set.

\begin{figure*}[t]
\centering
\includegraphics[width=0.99\textwidth]{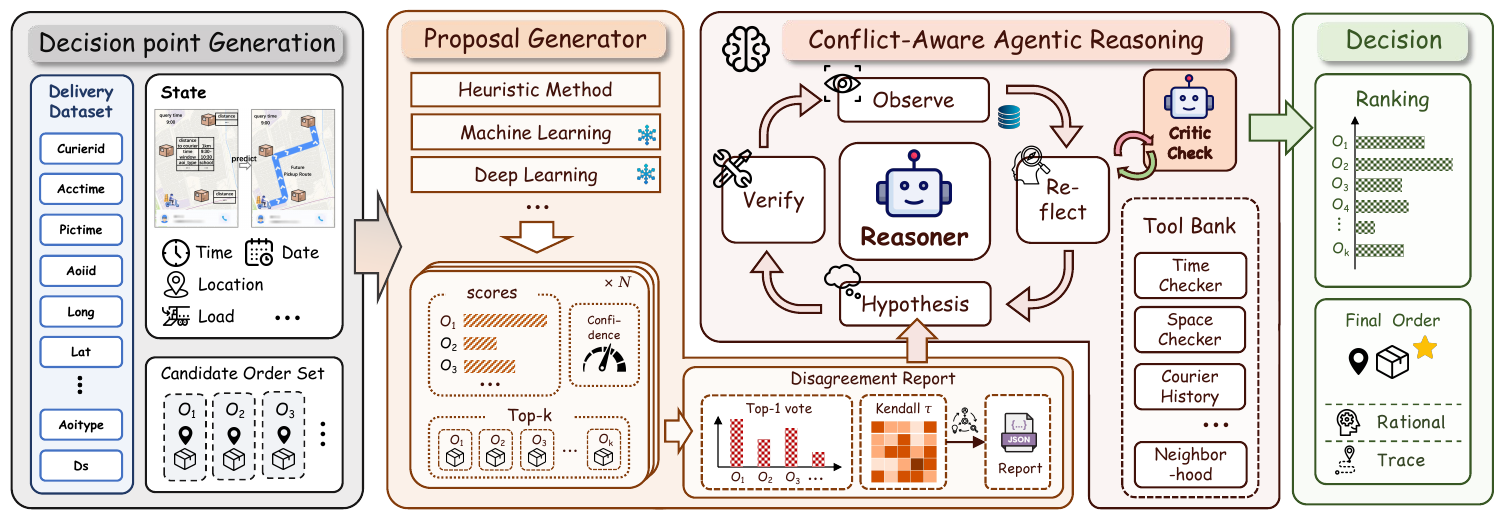}
\caption{Overall architecture of ORBITER.}
\label{fig:framework}
\end{figure*}

\section{Preliminaries}

\begin{definition}[\textbf{Location}]
\label{def:location}
In last-mile delivery, a location $p\in\mathcal{L}$ is a geographically distributed pickup or drop-off stop, represented by its coordinates and area of interest (AOI).
An order $o$ is associated with location $p(o)$, acceptance time $t_{\mathrm{acc}}(o)$, deadline $t_{\mathrm{ddl}}(o)$, and observed completion time $t_{\mathrm{cmp}}(o)$.
\end{definition}

\begin{definition}[\textbf{Courier Trajectory}]
\label{def:courier-trajectory}
Let $\mathcal{U}$ be the set of couriers.
For courier $u\in\mathcal{U}$ over a service period, the trajectory is the time-ordered sequence
\begin{equation}
\label{eq:user-trajectory}
\mathcal{T}_u=\big((p_1,\tau_1),(p_2,\tau_2),\ldots,(p_n,\tau_n)\big),
\end{equation}
where $p_i$ is the $i$-th location served and $\tau_i$ is the corresponding service time.
The trajectory records the realized route, but not the outstanding orders from which each next stop was chosen.
\end{definition}

\begin{definition}[\textbf{Decision Point}]
\label{def:decision-point}
Consider courier $u$ immediately after completing service at $(p_t,\tau_t)$.
The orders accepted by $\tau_t$ but not yet served form the candidate set $\mathcal{C}_t=\{o\mid t_{\mathrm{acc}}(o)\le\tau_t<t_{\mathrm{cmp}}(o)\}$.
Let $y_t$ denote the order served at the next trajectory record.
When $|\mathcal{C}_t|\ge2$ and $y_t\in\mathcal{C}_t$, we call this choice opportunity a \emph{decision point}, represented by
\begin{equation}
\label{eq:decision-state}
s_t=\bigl(u,p_t,\tau_t,\mathcal{T}_u^{<t},\mathcal{C}_t\bigr),
\end{equation}
where $\mathcal{T}_u^{<t}$ is the prefix of $\mathcal{T}_u$ completed before $\tau_t$.
Each $o\in\mathcal{C}_t$ is described by attributes $\mathbf{x}_t(o)$ observable at that time, including its location and temporal constraints.
Anchoring the state at $\tau_t$ makes each decision point reproducible from operational logs while preserving the information available before the next service event.
\end{definition}

\begin{problem}
We formulate next-order decision making at the level of individual service steps.
At decision point $s_t$, the agent observes an inference-safe state $\Phi(s_t)$ containing information available by $\tau_t$, with candidates presented independently of their future service order.
The decision agent $\pi$ returns a candidate ranking and commits to its first element:
\begin{equation}
\label{eq:agent-decision}
\hat\rho_t=\pi\!\left(\Phi(s_t)\right),
\qquad \hat y_t=\hat\rho_t(1).
\end{equation}

The objective is to place $y_t$ as high as possible in $\hat\rho_t$, thereby modeling a courier's service process as a sequence of step-level agent decisions.
\end{problem}

\section{Methodology}

Figure~\ref{fig:framework} gives an overview of ORBITER, which proceeds in four stages. First, \emph{Decision Point Generation} reconstructs an inference-safe candidate set at each service event from spatiotemporal order logs. Next, \emph{Heterogeneous Proposal Generator} ranks these candidates with fixed proposers of different inductive biases and summarizes their main disagreements in a structured report. Then, \emph{Conflict-Aware Agentic Reasoning} examines the resulting hypotheses with task-specific tools, while an independent critic reviews the conclusion. Finally, \emph{Evidence-Grounded Decision} determines the next order, retains the leading alternatives, and completes the ranking with the proposer outputs while preserving the accompanying decision record.

\subsection{Decision Point Generation}

Spatiotemporal order logs record the realized service sequence, but not the choice faced by the courier at each step. ORBITER reconstructs these choices and removes future-completion cues through leakage control.

\textbf{Decision reconstruction.}
To recover this missing decision context, ORBITER treats each service transition as a decision point with an explicit candidate set, processing each courier-day in completion order.
After the event $(p_t,\tau_t)$, the next completed order becomes $y_t$, and the orders visible at $\tau_t$ form $\mathcal{C}_t$ as specified in Definition~3.
States with $|\mathcal{C}_t|<2$ contain no meaningful choice, while states with $y_t\notin\mathcal{C}_t$ correspond to an order that was unavailable when the preceding service ended; both are excluded.
Writing $\mathcal{J}_u$ for the retained indices of courier $u$, the supervision set is
\begin{equation}
\label{eq:decision-dataset}
\begin{aligned}
\mathcal{Z}&=\bigl\{\bigl(\Phi(s_t),y_t\bigr)\bigm| u\in\mathcal{U},\ t\in\mathcal{J}_u\bigr\},\\
\mathcal{J}_u&=\bigl\{t\bigm| |\mathcal{C}_t|\ge2,\ y_t\in\mathcal{C}_t\bigr\}.
\end{aligned}
\end{equation}

Every label in $\mathcal{Z}$ is thus conditioned on the alternatives that were live at the corresponding service transition.
To make those alternatives comparable, each candidate $o\in\mathcal{C}_t$ is described by
\begin{equation}
\label{eq:candidate-quantities}
\begin{aligned}
w_t(o) &= \tau_t-t_{\mathrm{acc}}(o),\\
\ell_t(o) &= t_{\mathrm{ddl}}(o)-\tau_t,\\
d_t(o) &= \mathrm{dist}\bigl(p_t,p(o)\bigr),
\end{aligned}
\end{equation}
where $w_t(o)$ is the waiting time since acceptance, $\ell_t(o)$ the remaining slack to the deadline, and $d_t(o)$ the courier--order distance.
Together with location, AOI, and time-window attributes, these give every candidate a row $\mathbf{x}_t(o)\in\mathbb{R}^{d_x}$ with $d_x=8$; stacking the rows over $\mathcal{C}_t$ yields the candidate matrix $\mathbf{X}_t\in\mathbb{R}^{n_t\times d_x}$, $n_t=|\mathcal{C}_t|$, that every component receives.

\textbf{Leakage control.}
Benchmarks built from logs can reward train--test overlap rather than modeling, as shown for next-location prediction~\cite{luca2023overlap}.
The reconstructed candidate list inherits future completion order, so row position alone would reveal $y_t$.
We therefore sample a permutation $\sigma_t$ of $[n_t]$ from an episode-specific deterministic seed and apply it jointly to candidate identifiers, feature rows, and proposer inputs.
Consistent with the definition of $\Phi(s_t)$ in Section~3, the agent observes neither $y_t$ nor any future-route field; these variables are retained only as supervision.
Finally, $\mathcal{Z}$ is partitioned chronologically by service day so that all test days follow the training days.

\subsection{Heterogeneous Proposal Generator}
Given the reconstructed decision points, ORBITER obtains rankings from fixed proposers with different inductive
biases and organizes their disagreements for subsequent reasoning. It first generates heterogeneous proposals and then
summarizes them in a structured disagreement report.

\textbf{Heterogeneous proposers.}
LLMs can fall short when a decision requires exploring several plausible alternatives~\cite{yao2023tree}.
ORBITER therefore uses pretrained proposers to identify model-supported candidates before evidence collection, while preserving their competing views of the decision.

The proposers span three distinct inductive biases: heuristics, statistical machine learning, and deep learning.
For proposer $m$, let $\mathbf{z}_t^m\in\mathbb{R}^{n_t}$ denote its candidate scores, which order $\mathcal{C}_t$ by
\begin{equation}
\label{eq:proposal-ranking}
\begin{aligned}
r_{t,i}^{m}&=1+\bigl|\{\,j\in[n_t]\mid z_{t,j}^{m}>z_{t,i}^{m}\,\}\bigr|,\\
c_t^{m}&=\operatorname*{arg\,min}_{i\in[n_t]}\;r_{t,i}^{m},
\end{aligned}
\end{equation}
so that rank one marks the order proposer $m$ favors.
Keeping the whole rank vector rather than the vote alone preserves near ties and consistently supported alternatives.

Because rule scores, probabilities, and model logits are not directly comparable, each nonconstant score vector is min--max normalized and rescaled to a distribution $\mathbf{p}_t^m$, whose concentration is
\begin{equation}
\label{eq:proposal-confidence}
\kappa_t^m=1-\frac{H(\mathbf{p}_t^m)}{\log n_t},
\qquad
H(\mathbf{p})=-\sum_i p_i\log p_i.
\end{equation}

We set $\kappa_t^m=0$ for a constant score vector.
Each proposer thus contributes both a preference order over $\mathcal{C}_t$ and a measure of how sharply it separates its top choice.

\textbf{Disagreement report.}
Raw proposer rankings do not tell the agent which alternatives merit comparison or what drives the conflict.
Following structured prompting~\cite{beurerkellner2023prompting,besta2024graph}, ORBITER converts them into a disagreement report $\mathcal{D}_t$ containing candidate hypotheses and focused comparisons.

$\mathcal{H}_t^0$ begins with the distinct Top-1 proposals.
To retain alternatives repeatedly placed near the top, ORBITER supplements this set by reciprocal-rank support:
\begin{equation}
\label{eq:proposer-rrf}
\operatorname{RRF}_t(i)=\sum_{m\in[M]}\frac{1}{r_{t,i}^{m}},
\end{equation}
Candidates with the largest values are then added to $\mathcal{H}_t^0$.
The Top-1 choice of the most contextually reliable proposer then serves as a reference:
\begin{equation}
\label{eq:reference-proposal}
m_t^\star=\operatorname*{arg\,max}_{m\in[M]}\alpha_t^m,
\qquad
\hat y_t^0=c_t^{m_t^\star},
\end{equation}
where $\alpha_t^m$ is estimated from training data.
This reference starts the comparison but does not determine its outcome.

The report describes the resulting challenge through Top-1 votes and full rankings:
\begin{equation}
\label{eq:vote-disagreement}
q_t(c)=\frac{1}{M}\sum_{m\in[M]}\mathbb{I}\bigl[c_t^{m}=c\bigr],
\qquad
V_t=\frac{H(\mathbf{q}_t)}{\log M}.
\end{equation}
Here, $V_t$ measures how widely the panel splits.
Pairwise Kendall correlations show whether that split extends beyond Top-1, while
\begin{equation}
\label{eq:rank-span}
R_t(i)=\max_{m\in[M]}r_{t,i}^{m}-\min_{m\in[M]}r_{t,i}^{m}.
\end{equation}
locates candidate-specific disputes.
Together with the candidate attributes, these statistics expose the principal trade-off between the reference and its leading rival and formulate the question tested in the next stage.

\subsection{Conflict-Aware Agentic Reasoning}

The disagreement report identifies the competing candidates but not the evidence that settles their dispute.
ORBITER follows the reasoning--action alternation of ReAct~\cite{yao2023react}, while tying each tool call to one unresolved comparison.
The resulting loop tests that comparison before an independent critic examines the proposed conclusion.

\textbf{Hypothesis state.}
Each candidate $c\in\mathcal{H}_t^0$ induces the testable claim that $c$ should be served next.
At step $k$, the reasoning state is
\begin{equation}
\label{eq:agent-state}
\Omega_t^k
=
\bigl(
\mathcal{H}_t^k,\,
\mathcal{Q}_t^k,\,
\mathcal{E}_t^k
\bigr),
\end{equation}
where $\mathcal{H}_t^k$ stores the claims and their status, $\mathcal{Q}_t^k$ the unresolved comparisons, and $\mathcal{E}_t^k$ the evidence collected so far.
Candidate identities remain fixed; observations mark their claims as supported, contradicted, or contested.

\textbf{Evidence collection.}
Given this state, the report, and tool set $\mathcal{F}$, the controller chooses
\begin{equation}
\label{eq:agent-action}
a_t^k
\sim
\pi_{\mathrm{LLM}}\!\left(
\cdot\mid\Omega_t^k,\mathcal{D}_t,\mathcal{F}
\right),
\end{equation}
which invokes a tool, formulates the next comparison, or submits a proposal in $\mathcal{H}_t^0$.
Every query names the candidate, its rival, and the outcome that would change their comparison.
When $a_t^k$ invokes tool $f_t^k$ on candidate $c$ and rival $\bar c$, evidence is acquired by
\begin{equation}
\label{eq:evidence-acquisition}
\begin{aligned}
e_t^k
&=f_t^k\!\left(
\Phi(s_t),\,c,\,\bar c
\right),\\[2pt]
\mathcal{E}_t^{k+1}
&=
\begin{cases}
\mathcal{E}_t^k\cup\{e_t^k\},
& e_t^k\ \text{is admissible},\\
\mathcal{E}_t^k,
& \text{otherwise}.
\end{cases}
\end{aligned}
\end{equation}
Admissibility requires a time-safe, candidate-directed result with traceable provenance.
Spatiotemporal tools test the current trade-off, historical tools retrieve comparable decisions, and robustness tools probe plausible perturbations.

Let $\mathcal{E}_{+}=\mathcal{E}_t^{k,+}(c)$ and $\mathcal{E}_{-}=\mathcal{E}_t^{k,-}(c)$ denote the admissible evidence supporting and opposing $c$.
The hypothesis status is updated as
\begin{equation}
\label{eq:hypothesis-update}
h_t^{k+1}(c)
=
\begin{cases}
\text{supported},
& \mathcal{E}_{+}\ne\varnothing,\ 
  \mathcal{E}_{-}=\varnothing,\\
\text{contradicted},
& \mathcal{E}_{+}=\varnothing,\ 
  \mathcal{E}_{-}\ne\varnothing,\\
\text{contested},
& \mathcal{E}_{+}\ne\varnothing,\ 
  \mathcal{E}_{-}\ne\varnothing,\\
\text{untested},
& \text{otherwise}.
\end{cases}
\end{equation}

\textbf{Critic.}
The critic reads the proposal and evidence independently and may approve, reject, or request another observation without proposing a candidate itself.
It may approve a proposal only if the selected candidate lies in $\mathcal{H}_t^0$, cites admissible supporting evidence, and addresses its strongest rival.
Otherwise, the critic returns the missing comparison to the loop, subject to the evidence budget.

%

\subsection{Evidence-Grounded Decision}

Let $\mathcal{A}_{\mathrm{LLM}}$ denote the complete conflict-aware agent, including the hypothesis-verification loop and critic feedback, and $\mathcal{R}_t$ its evidence record.
The resulting agentic decision is
\begin{equation}
\label{eq:final-decision}
\bigl(\hat y_t,\mathcal{R}_t\bigr)
=
\mathcal{A}_{\mathrm{LLM}}\!\left(
\Phi(s_t),\mathcal{D}_t;\mathcal{F}
\right),
\qquad
\hat y_t\in\mathcal{H}_t^0.
\end{equation}
The first component supplies Top-1; proposer outputs only complete the remaining ranking.

Let $\widetilde z_{t,i}^m$ denote candidate $i$'s min--max normalized score from proposer $m$.
Using the concentration in Equation~\eqref{eq:proposal-confidence}, its fused score is
\begin{equation}
\label{eq:fused-ranking}
g_t(i)
=
\frac{
\sum_{m=1}^{M}
\kappa_t^m\widetilde z_{t,i}^m
}{
\sum_{m=1}^{M}\kappa_t^m
}.
\end{equation}
Sharper proposer distributions receive greater weight; if all concentration values are zero, we use the unweighted mean.

ORBITER retains the agent decision and two leading rivals from the hypothesis set as its Top-3.
The rivals follow $\hat y_t$ in decreasing order of $g_t(i)$, and the same score ranks all remaining candidates.
This keeps the principal conflict visible in the returned ranking.
The record $\mathcal{R}_t$ contains the supporting evidence, treatment of the leading counterevidence, and critic review behind the decision. 

\section{Experiments}

\subsection{Experimental Setup}
\noindent\textbf{Datasets.}
We use four city subsets of \textbf{LaDe}-P \cite{wu2024lade}: \textbf{Shanghai}, \textbf{Chongqing}, \textbf{Jilin}, and \textbf{Yantai}.
Released by Cainiao, LaDe contains 10.677 million packages served by 21,000 couriers over six months.
Shanghai and Chongqing are high-volume cities, with Chongqing featuring a more complex road network; Jilin and Yantai represent medium- and small-sized cities.
The pickup records provide locations, service time windows, event times, AOI attributes, and courier information. 

\begin{table}[h]
\centering
\small
\begin{tabular*}{\columnwidth}{@{\extracolsep{\fill}}cccc}
\toprule
\textbf{City} & \textbf{Packages} & \textbf{Couriers} & \textbf{AvgPackage} \\
\midrule
Shanghai  & 1,450k & 4,502 & 15.0 \\
Chongqing & 1,172k & 2,982 & 14.0 \\
Yantai    & 1,146k & 2,593 & 16.0 \\
Jilin     & 261k   & 665   & 13.8 \\
\bottomrule
\end{tabular*}
\caption{Statistics of the LaDe-P subsets.}
\label{tab:dataset}
\end{table}

\noindent\textbf{Baselines and Evaluation.}
We compare ORBITER with thirteen baselines from four categories: heuristic methods (Distance Greedy, Deadline Greedy, and Weighted Rule), machine learning models (LightGBM \cite{ke2017lightgbm}, XGBoost \cite{chen2016xgboost}, and Random Forest), deep learning models (DeepRoute \cite{wen2021deeproute}, Graph2Route \cite{wen2022graph2route}, DRL4Route \cite{mao2023drl4route}, and MRGRP \cite{liu2025mrgrp}), and LLM-based methods (LLM-Mob \cite{wang2023llmmob}, LLM-Move \cite{feng2024llmmove}, and AgentMove \cite{feng2025agentmove}). We report ACC@1, ACC@3, and mean reciprocal rank (MRR) under the same data splits and candidate sets. 

\subsection{Main Results}


\begin{table*}[t]
\centering
\footnotesize
\renewcommand{\arraystretch}{1.0}
\setlength{\tabcolsep}{4pt}

\begin{tabular}{l*{4}{ccc}}
\toprule
\textbf{Model}
& \multicolumn{3}{c}{\textbf{Shanghai}}
& \multicolumn{3}{c}{\textbf{Jilin}}
& \multicolumn{3}{c}{\textbf{Chongqing}}
& \multicolumn{3}{c}{\textbf{Yantai}} \\

\cmidrule(lr){2-4}
\cmidrule(lr){5-7}
\cmidrule(lr){8-10}
\cmidrule(lr){11-13}

& ACC@1 & ACC@3 & MRR
& ACC@1 & ACC@3 & MRR
& ACC@1 & ACC@3 & MRR
& ACC@1 & ACC@3 & MRR \\
\midrule

Distance Greedy
& 0.250 & 0.575 & 0.452
& 0.255 & 0.570 & 0.465
& 0.280 & 0.600 & 0.478
& 0.280 & 0.610 & 0.483 \\

Deadline Greedy
& 0.205 & 0.505 & 0.414
& 0.185 & 0.430 & 0.383
& 0.220 & 0.565 & 0.440
& 0.210 & 0.535 & 0.414 \\

Weighted Rule
& 0.435 & 0.755 & 0.628
& 0.350 & 0.655 & 0.543
& 0.470 & 0.745 & 0.636
& 0.480 & 0.760 & 0.640 \\

\midrule

LightGBM
& 0.475 & 0.735 & 0.644
& 0.480 & 0.755 & 0.646
& 0.505 & 0.785 & 0.665
& 0.450 & 0.790 & 0.644 \\

XGBoost
& 0.500 & 0.760 & 0.660
& \underline{0.485} & 0.765 & \underline{0.655}
& 0.525 & 0.790 & 0.680
& 0.450 & 0.795 & 0.642 \\

Random Forest
& 0.450 & 0.785 & 0.639
& 0.405 & 0.740 & 0.603
& 0.480 & \underline{0.815} & 0.663
& 0.460 & 0.795 & 0.645 \\

\midrule

DeepRoute
& 0.490 & 0.810 & 0.664
& 0.455 & 0.775 & 0.636
& 0.505 & 0.785 & 0.661
& 0.515 & \underline{0.840} & 0.686 \\

Graph2Route
& 0.500 & 0.795 & 0.666
& 0.430 & 0.740 & 0.617
& 0.535 & 0.790 & 0.682
& 0.510 & 0.820 & 0.681 \\

DRL4Route
& 0.495 & 0.815 & 0.667
& 0.460 & 0.780 & 0.641
& 0.520 & 0.800 & 0.666
& \underline{0.525} & \textbf{0.845} & \underline{0.691} \\

MRGRP
& \underline{0.505} & \textbf{0.830} & \underline{0.669}
& 0.455 & \underline{0.785} & 0.632
& \underline{0.550} & 0.805 & \underline{0.692}
& 0.495 & 0.810 & 0.665 \\

\midrule

LLM-Mob
& 0.340 & 0.680 & 0.521
& 0.285 & 0.630 & 0.463
& 0.375 & 0.695 & 0.541
& 0.325 & 0.665 & 0.503 \\

LLM-Move
& 0.355 & 0.600 & 0.503
& 0.285 & 0.560 & 0.437
& 0.325 & 0.625 & 0.482
& 0.335 & 0.595 & 0.480 \\

AgentMove
& 0.350 & 0.640 & 0.508
& 0.290 & 0.535 & 0.426
& 0.300 & 0.555 & 0.450
& 0.285 & 0.580 & 0.446 \\

\midrule

\textbf{ORBITER}
& \textbf{0.535} & \underline{0.820} & \textbf{0.689}
& \textbf{0.515} & \textbf{0.800} & \textbf{0.677}
& \textbf{0.590} & \textbf{0.835} & \textbf{0.718}
& \textbf{0.550} & \underline{0.840} & \textbf{0.704} \\

\bottomrule
\end{tabular}
\caption{Main results on the four city subsets of LaDe-P. All LLM-based methods use DeepSeek-V4-Flash as the backbone.}
\label{tab:overall-performance}
\end{table*}

Table~\ref{tab:overall-performance} compares all methods across four LaDe-P city subsets using the same candidate sets. ACC@1 measures the committed next-order decision, while MRR captures the rank of the true order. The following observations are made.

First, \emph{ORBITER performs consistently well across cities.} It achieves the highest ACC@1 and MRR in every city, outperforming the strongest baseline by 9.2\% and 4.9\% on average, respectively. By checking competing proposals against task-specific evidence, the agent can revise an incorrect top choice without discarding the ranking prior supplied by the trained predictors. Its advantage therefore appears in both next-order decision accuracy and the rank of the true order, rather than in a single city or metric. Meanwhile, \emph{ORBITER makes effective use of heterogeneous proposals.} Among the baselines, MRGRP leads ACC@1 in Shanghai and Chongqing, while XGBoost and DRL4Route lead in Jilin and Yantai. These shifts indicate that different model families capture complementary decision cues. Rather than committing to one of them, ORBITER retains their competing views as proposals: the disagreement report narrows the comparison to model-supported candidates, while targeted evidence resolves the remaining conflict. It consequently ranks first in ACC@1 and MRR across all four cities even as the strongest individual predictor changes. Finally, \emph{ORBITER makes more effective use of LLM reasoning.} Across the four cities, the three LLM baselines achieve an average ACC@1 of 32.1\%, 41.4\% lower than ORBITER and broadly comparable to the heuristic methods. Designed for mobility prediction, these methods rely heavily on recurring locations and trajectory history. Delivery decision making instead requires choosing from a changing set of orders whose distance, waiting time, and deadlines may conflict. ORBITER preserves the task-specific ranking priors of its proposers and asks the LLM to resolve disagreements among their leading candidates, rather than choose the next order from scratch.

\begin{figure}[h]
    \centering
    \hspace*{-8pt}%
    \includegraphics[width=1\linewidth]{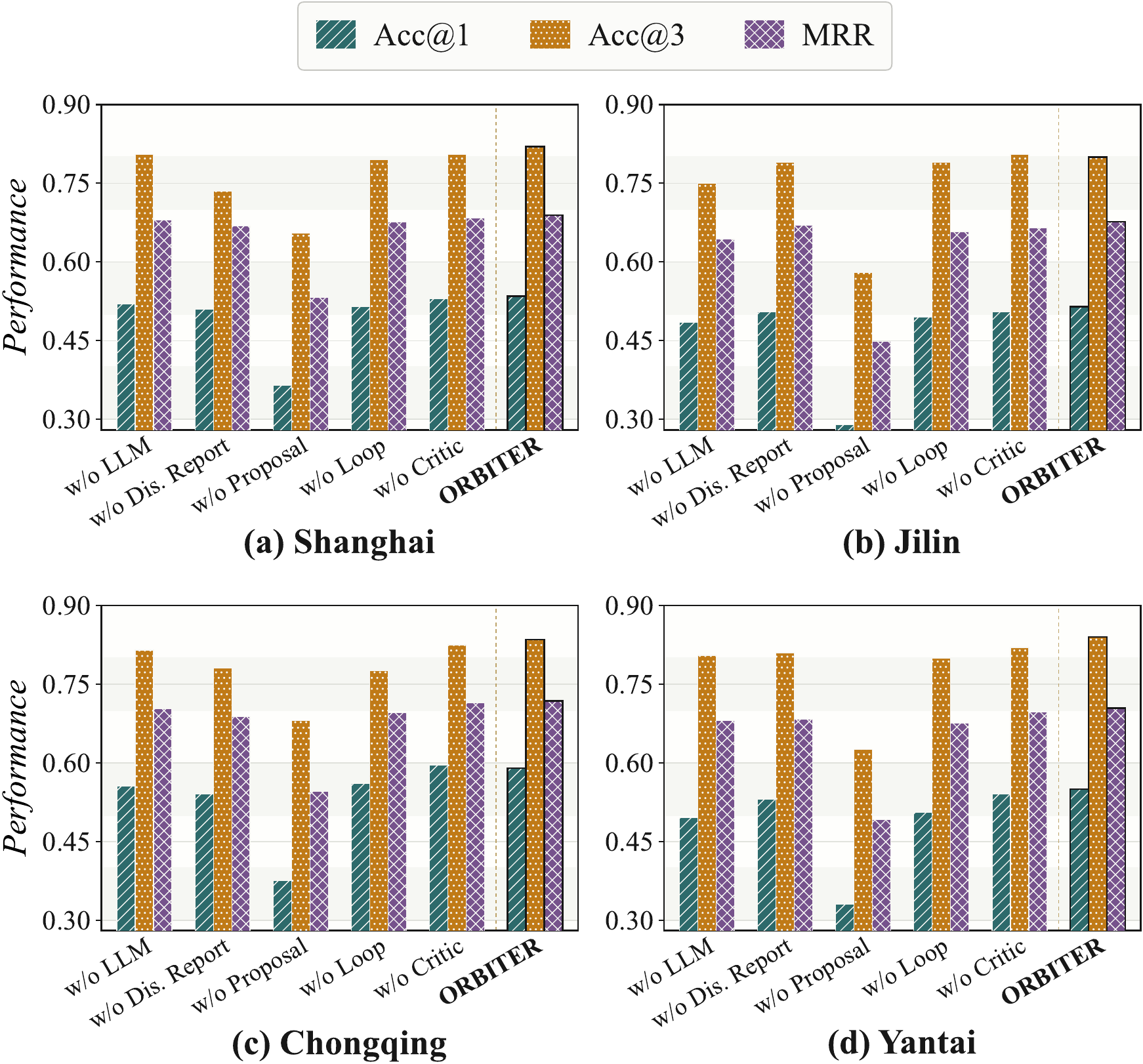}
    \caption{Ablation studies.}
    \vspace{-0.55cm}
    \label{fig:Ablation}
\end{figure}

\subsection{Ablation Studies}

We study five controlled variants.
\emph{w/o Proposal Models} uses only the candidate state tools.
\emph{w/o LLM} replaces reasoning and evidence collection with confidence-weighted proposer fusion.
\emph{w/o Disagreement Report} gives the agent raw proposer outputs.
\emph{w/o Reasoning Loop} replaces the hypothesis-verification loop and evidence revision with a single-pass LLM decision.
\emph{w/o Critic} removes the independent critic.

Figure~\ref{fig:Ablation} reports the relative losses from each component removal. (1)~\emph{The heterogeneous proposers provide ORBITER's primary task guidance.} Removing proposer guidance lowers mean ACC@1 by 37.98\%, bringing performance close to the LLM-based baselines. Candidate states and tools alone do not reliably identify the next order from the full set, making proposer rankings the basis for subsequent reasoning.
(2)~\emph{Evidence-based agent reasoning adds value beyond proposal fusion.} Replacing reasoning and evidence collection with confidence-weighted fusion lowers mean ACC@1 by 6.14\%. Heterogeneous proposals identify plausible candidates, but score fusion alone cannot resolve the conflicts among them.
(3)~\emph{The disagreement report directs the agent to the relevant conflicts.} Without the structured report, mean ACC@1 falls by 4.68\%. Encoding split votes and rank spans makes proposer-supported conflicts explicit, rather than leaving the agent to locate them in redundant raw outputs.
(4)~\emph{The hypothesis-verification loop adapts evidence collection to unresolved conflicts.} Fixed evidence and one-shot inference lower mean ACC@1 by 4.10\%. One observation may rule out a candidate without separating the remaining hypotheses, so evidence collection must follow the conflict left unresolved.
(5)~\emph{The independent critic provides a final corrective check.} Removing independent review lowers the three metrics by 1.06\% on average. Its effect varies by city, indicating modest but measurable corrections to the final decision.
\subsection{Sensitivity to the backbone LLM}

\begin{figure}[t]
    \centering
    \hspace*{-8pt}%
    \includegraphics[width=1\linewidth]{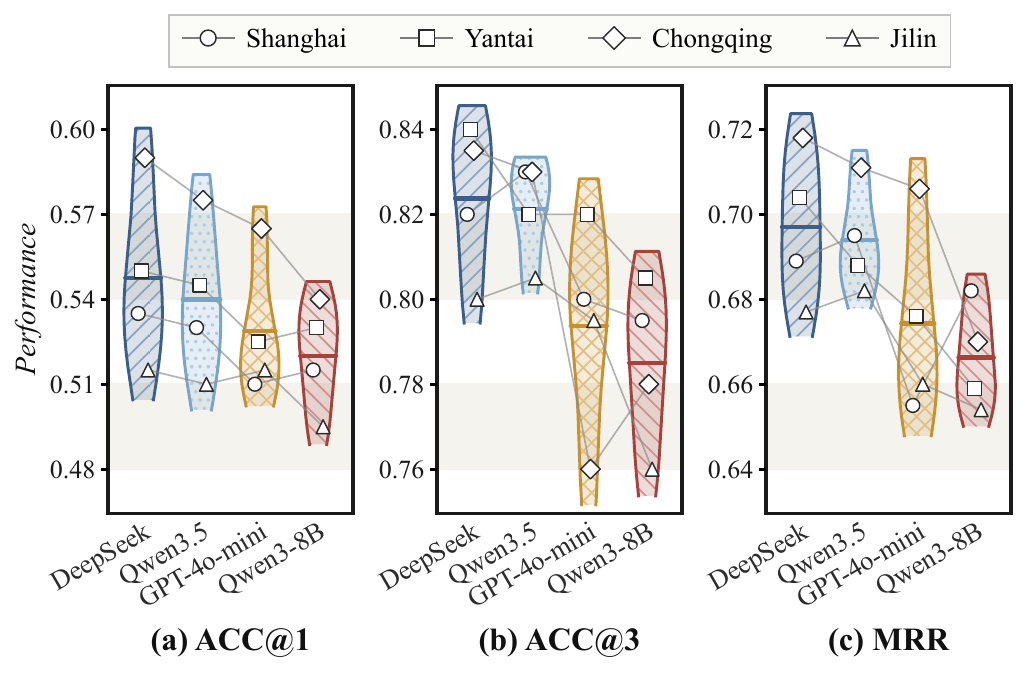}
    \caption{Performance across backbone LLMs.}
    \label{fig:backbone_sensitivity}
\end{figure}

To assess ORBITER's sensitivity to its backbone LLM, we evaluate Qwen3.5-Flash,\footnote{\url{https://qwen.ai/blog?id=qwen3.5}} GPT-4o-mini,\footnote{\url{https://developers.openai.com/api/docs/models/gpt-4o-mini}} and Qwen3-8B \cite{qwen2025qwen3} alongside DeepSeek-V4-Flash \cite{deepseek2026v4} across the four cities, keeping the proposers, tools, and reasoning budget fixed.
Figure~\ref{fig:backbone_sensitivity} shows that DeepSeek performs best, while Qwen3.5 trails by only 1.4\% in mean ACC@1.
GPT-4o-mini and Qwen3-8B lower this metric by 3.4\% and 5.0\%; ACC@3 and MRR follow the same trend.
DeepSeek targets agentic tool use, and Qwen3.5 supports tool-oriented workflows, helping both revise hypotheses across calls.
GPT-4o-mini lacks documented agent-specific optimization, whereas Qwen3-8B is limited by its parameter scale.
Across all backbones, proposer-supplied candidates and ranking priors keep the decision space bounded, preserving performance when LLM reasoning is weaker.

\subsection{Case Study}

We examine a Shanghai decision point to illustrate how ORBITER moves from proposer disagreement to an evidence-grounded decision.

\begin{figure}[t]
    \centering
    \includegraphics[width=1\linewidth]{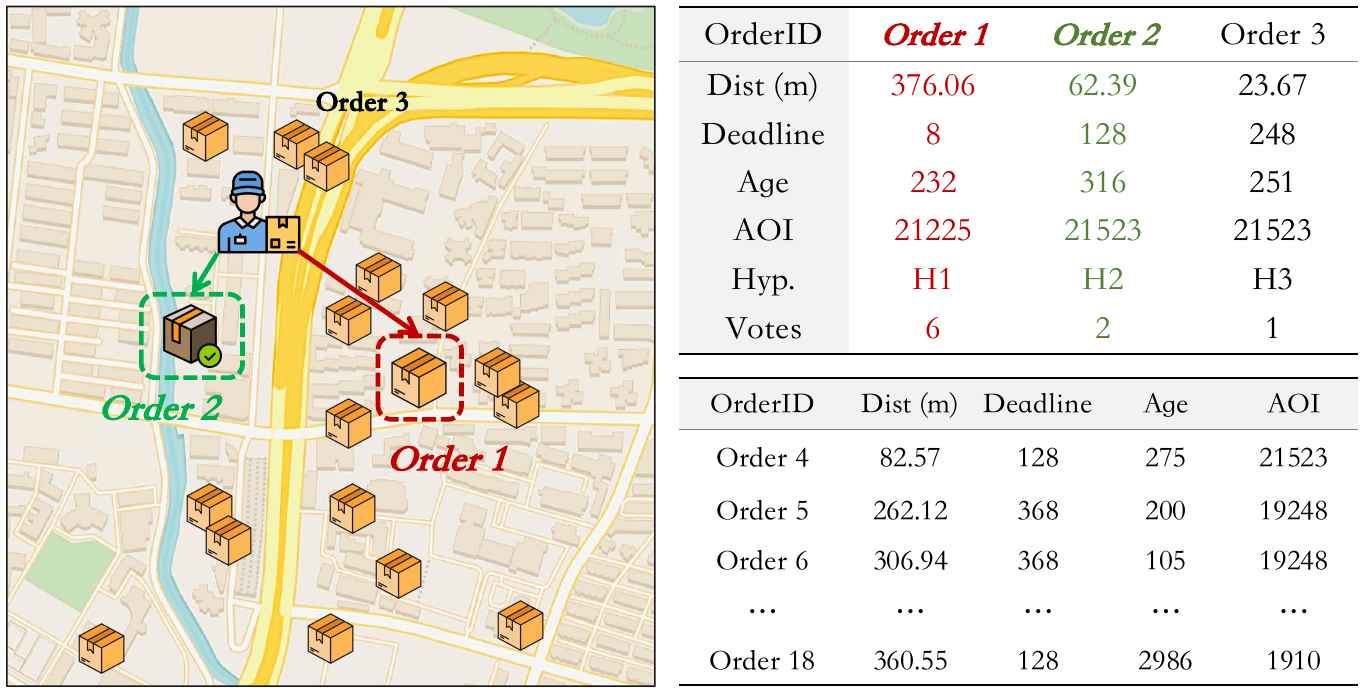}
    \caption{Conflicting cues among candidate orders.
}
    \label{fig:case_map}
\end{figure}

\noindent\textbf{Decision context.}
After completing a pickup, the courier faces 18 visible orders, as shown in Figure~\ref{fig:case_map}. The distance rule favors Order~3, only 23.67~m away, whereas the deadline rule favors Order~1, 376.06~m away with 8~min of slack. The courier actually serves Order~2 next: it is 62.39~m away, with 128~min of slack and several orders in the same AOI.

\begin{figure}[h]
    \centering
    \includegraphics[width=0.9\linewidth]{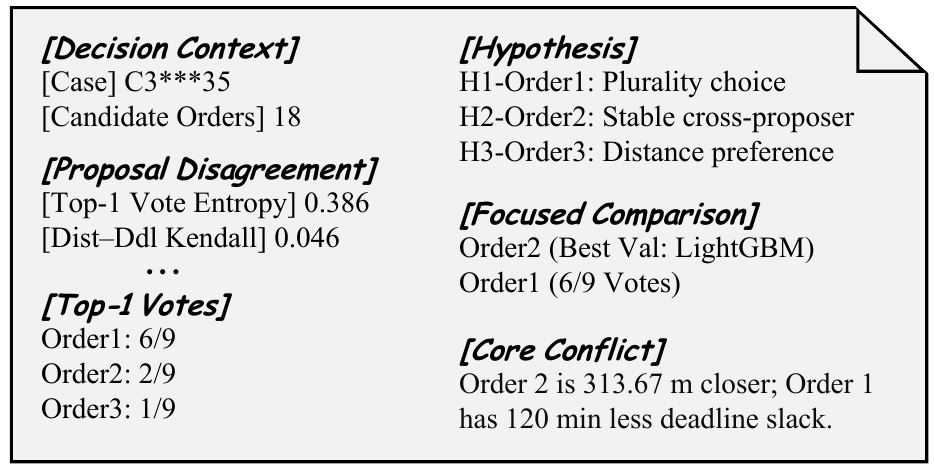}
    \caption{Example of a Structured Disagreement Report.}
    \label{fig:report}
\end{figure}

\noindent\textbf{Disagreement report.}
The nine proposers split 6:2:1 on their Top-1 choices. The report in Figure~\ref{fig:report} retains three hypotheses and reveals what the vote count misses: Order~2 receives only two Top-1 votes, but all nine proposers rank it in the Top-3.
Confidence-weighted fusion nevertheless selects Order~1.

\noindent\textbf{Agentic reasoning.}
ORBITER treats deadline pressure and spatial clustering as competing hypotheses.
A deadline check confirms Order~1's urgency, supporting the premise shared by six proposers.
An AOI-density query then places Order~2 in a denser cluster of orders from the same AOI, introducing counterevidence.
To determine whether it can safely precede Order~1, the agent runs a counterfactual route check.
Serving Order~2 first adds only 12.68~m of detour and still reaches Order~1 in 3.62~min, before its deadline.
The deadline hypothesis therefore becomes contested: the urgency is real but does not require immediate service. After an independent review, the critic approves the proposal, and ORBITER selects Order~2, matching the courier's actual next order.
The case shows how targeted evidence can overturn a majority proposal while retaining an auditable decision record.

\section{Conclusion}
We present ORBITER, a conflict-aware agentic framework for next-order decision making in last-mile delivery. ORBITER generates decision points from spatiotemporal order logs, retaining the courier state and visible orders at each service event. Heterogeneous proposers rank these orders by different decision cues, while a structured disagreement report identifies competing hypotheses. Conflict-Aware Agentic Reasoning tests these hypotheses with task-specific evidence, and an independent critic reviews the proposed decision. ORBITER returns the next-order decision with an auditable evidence record. Experiments on LaDe-P data from four cities show that ORBITER achieves the best ACC@1 and MRR in every city.

\bibliography{aaai2027}

\end{document}